\documentclass[lettersize,journal]{IEEEtran}
\usepackage{amsmath,amsfonts}
\usepackage{algorithmic}
\usepackage{algorithm}
\usepackage{array}
\usepackage[caption=false,font=normalsize,labelfont=sf,textfont=sf]{subfig}
\usepackage{textcomp}
\usepackage{stfloats}
\usepackage{url}
\usepackage{verbatim}
\usepackage{graphicx}
\usepackage{cite}
\usepackage{epsfig}
\usepackage{graphicx}
\usepackage{xcolor}
\usepackage{caption}

\definecolor{orange}{rgb}{1.0, 0.5, 0.0}

\begin{document}

\title{Gesture2Path: Imitation Learning for \\ Gesture-aware Navigation}

\author{Catie Cuan$^{1, 2}$, Edward Lee$^{3}$, Emre Fisher$^{2}$, Anthony Francis$^{3}$, \\ Leila Takayama$^{3}$, Tingnan Zhang$^{3}$, Alexander Toshev$^{4, \dagger}$,  and S\"{o}ren Pirk$^{5, \dagger}$
\thanks{\footnotesize $^{1}$Department of Mechanical Engineering, Stanford University, Stanford, 94305, United States}%
\thanks{\footnotesize $^{2}$Everyday Robots, 100 Mayfield Ave, Palo Alto, CA 94043, USA}    
\thanks{\footnotesize $^{3}$Robotics at Google, 1600 Amphitheatre Parkway, Mountain View, CA 94304, USA}%
\thanks{\footnotesize $^{4}$Apple ML Research, Apple Park 1, Cupertino, CA 95014, USA}%
\thanks{\footnotesize $^{5}$Adobe Research, 345 Park Avenue, San Jose, CA 95110-2704, USA}
\thanks{\footnotesize $\dagger$ Work done while at Robotics at Google.}
}

\maketitle

\begin{abstract}
As robots increasingly enter human-centered environments, they must not only be able to navigate safely around humans, but also adhere to complex social norms. Humans often rely on non-verbal communication through gestures and facial expressions when navigating around other people, especially in densely occupied spaces. Consequently, robots also need to be able to interpret gestures as part of solving social navigation tasks. To this end, we present \textit{Gesture2Path}, a novel social navigation approach that combines image-based imitation learning with model-predictive control. Gestures are interpreted based on a neural network that operates on streams of images, while we use a state-of-the-art model predictive control algorithm to solve point-to-point navigation tasks. We deploy our method on real robots and showcase the effectiveness of our approach for the four gestures-navigation scenarios: left/right, follow me, and make a circle. Our experiments indicate that our method is able to successfully interpret complex human gestures and to use them as a signal to generate socially compliant trajectories for navigation tasks. We validated our method based on in-situ ratings of participants interacting with the robots. 
\end{abstract}

\section{INTRODUCTION}

Situated agents should not only navigate safely around people, but should also abide by social norms and respond to the full gamut of human behavior -- including  nonverbal communication such as body language, gestures, and facial expressions. A robot solving a navigation task must be able to interpret human behavior and to carefully adjust its actions to be socially compliant. We refer to this form of navigation task as \textit{Nonverbal Social Navigation}. For defining and generating social behavior, robotics research has recently expanded efforts toward understanding the importance of respecting personal space~\cite{1308100} and social dynamics~\cite{8036225,10.1145/2696454.2696463}, socially-acceptable behavior for approaching humans~\cite{huang2014}, navigation among groups of people~\cite{10.1007/978-3-030-50334-5_24}, the validation of socially acceptable policies~\cite{Pirk2022SocialNavProtocol}, and curating large datasets~\cite{Karnan2022}. The breadth of these research directions is a testament of the complexity of generating socially compliant agent behavior. 

\begin{figure}[th!]
\vspace{2mm}
\centering
\includegraphics[width=\columnwidth]{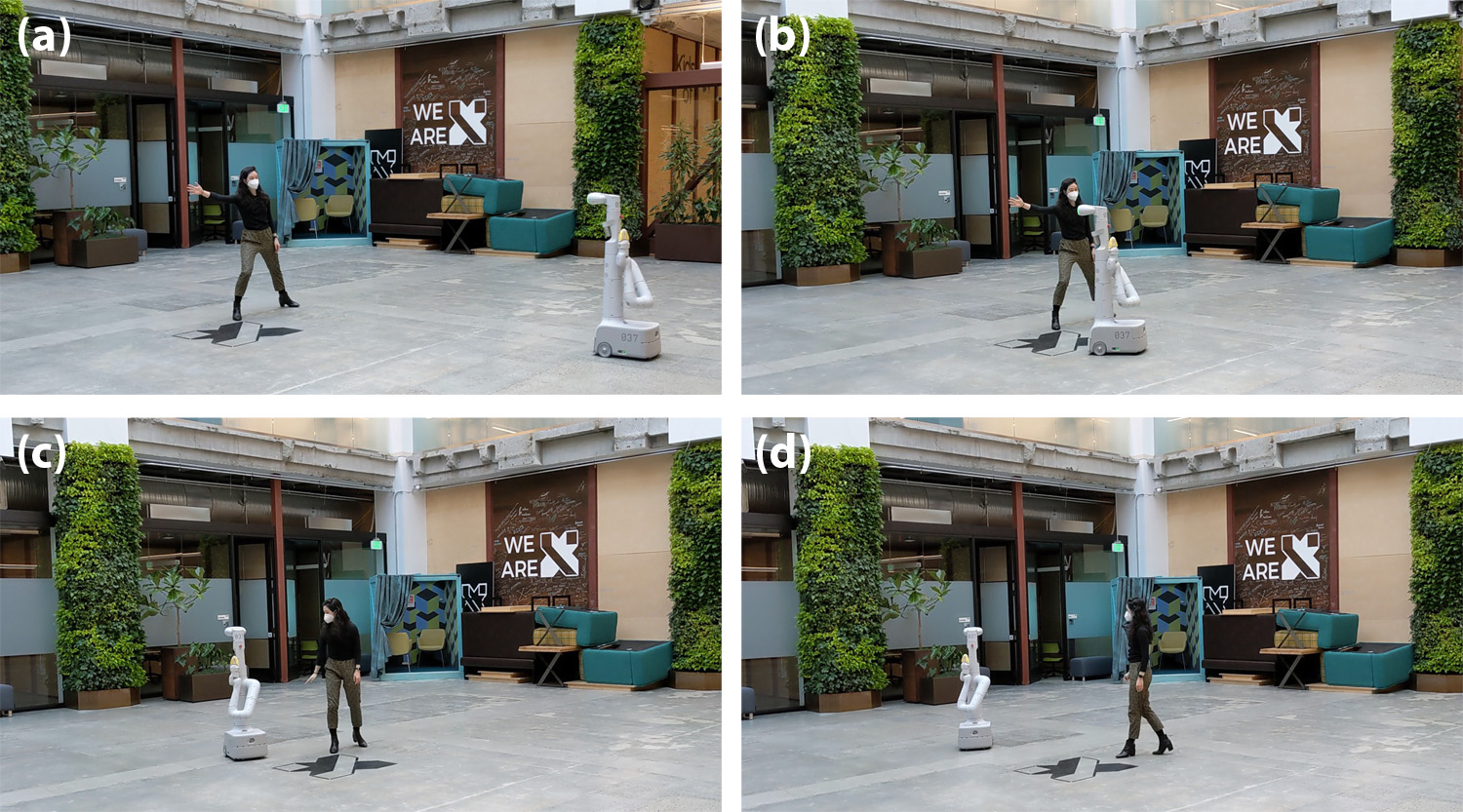}
\vspace{-5mm}
\caption{\small{\textit{Gesture2Path} is a novel social navigation policy that combines image-based imitation learning with model-predictive control to enable gesture-aware navigation. In this example we show our \textit{right} gesture policy: A robot at its start location begins navigating toward its goal (a). On its way, it encounters a person that indicates to the robot with a \textit{right} gesture to pass on their right (b). The robot interprets this gesture and drives around the person in the intended manner (c) to then continue its path towards its goal (d). The robot in use is from Everyday Robots.
}}
\vspace{-2mm}
\label{fig:gesture_teaser}
\end{figure}

\begin{figure*}[ht!]
\centering
\includegraphics[width=\linewidth]{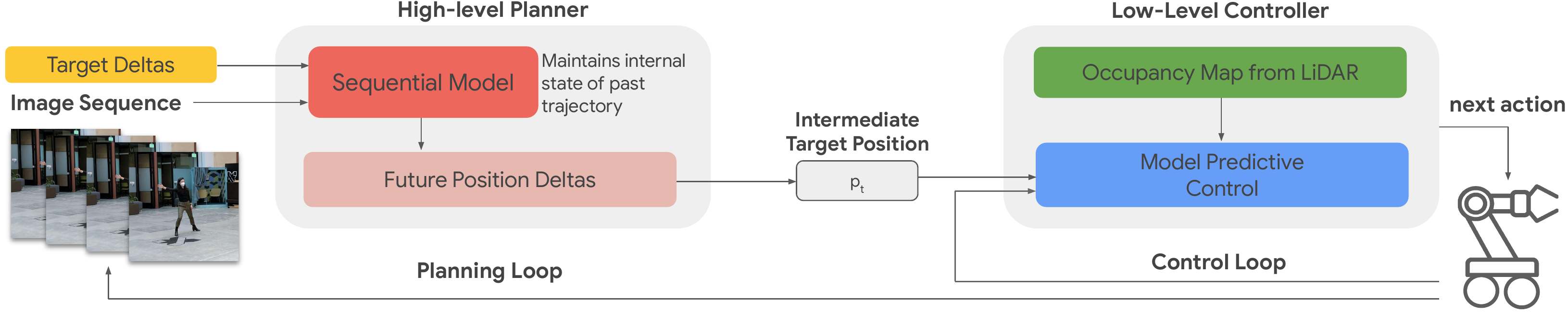}
\vspace{-6mm}
\caption{\small{
Overview of \textit{Gesture2Path}: A sequential neural network trained via imitation learning acts as a high-level planner to predict waypoint deltas from image sequences. The predicted waypoint deltas are used to compute a world space waypoint that serves as intermediate target position. MPC uses these intermediate target positions to compute linear and angular velocities to drive the robot.}}
\vspace{-5mm}
\label{fig:pipeline}
\end{figure*}

To solve navigation tasks, many existing approaches rely on point cloud data obtained from LiDAR scanners that provide real-time information about the environment, including dynamic objects such as humans. The captured point clouds are projected to 2-D occupancy maps. With these maps existing policy algorithms -- such as model predictive control (MPC) -- can efficiently solve complex navigation tasks with remarkable success. However, while point cloud data is a powerful sensor modality for generating navigation trajectories in the environment, it rarely provides the accuracy needed to interpret intricate human behavior. Conversely, while RGB sensors do not provide the depth information needed to solve navigation tasks, they produce high-resolution images that allow the capture of nuanced gestures and facial expressions. 

In this paper, we propose a novel gesture-aware navigation policy based on imitation learning and MPC. We train a sequential neural network that enables us to generate waypoints of navigation trajectories from a sequence of consecutive images. Our goal is to train this network so that it predicts waypoints that are socially compliant and adherent to the gestures of humans interacting with the robot. Once the network is trained, the predicted waypoints are sent to an MPC algorithm to control the robot. The sequential neural network serves as a high-level planner, while the MPC algorithm provides low-level control. This setup provides the benefits of both approaches: the sequential neural network allows us to obtain nuanced information of human gestures from sequences of images, while the MPC algorithm allows us to safely navigate the robot based on laser-scanned point clouds. 

To explore the effectiveness of our gesture-aware social navigation policy, we define the four gesture scenarios: Left/Right, Follow Me, and Make a Circle. For each scenario we define the gesture (e.g. pointing with the hand to the left) and the robot behavior (e.g. drive left) so each gesture maps to a specific robot behavior -- an example is shown in Fig.~\ref{fig:gesture_teaser}. We then collected a dataset of expert examples for each scenario by driving the robot with a human operator. A collected trajectory is defined by a sequence of images and a sequence of robot waypoints. We then train the sequential neural network on subsequences of images to predict subsequences of future waypoints. The MPC algorithm then computes the linear and angular velocities to drive from robots' current position to the predicted position of the sequential model. 

We validate our gesture-aware social navigation policy based on in-situ experience ratings of humans interacting with the robots. To obtain these ratings we follow a protocol for social navigation policies~\cite{Pirk2022SocialNavProtocol}. Each of our scenarios is defined by a gesture as well as the start and end positions of the trajectory that the robot is supposed to drive along, mirroring the training setup. Additionally, we define a questionnaire based on a five-level Likert scale for each scenario. Once a scenario is completed a participant is asked to provide their ratings for a specific scenario. We use this setup to compare our \textit{Gesture2Path} policy with a MPC policy as the baseline. 

In summary, our contributions are: 
(1) we introduce \textit{Gesture2Path}, a novel gesture-aware social navigation policy that combines a sequential neural network with MPC;
(2) we show that our policy is able to interpret gestures to generate socially-compliant behavior in gesture-navigation scenarios; 
(3) we define a canonical set of social navigation scenarios with gestures that can be replicated and tested; 
(4) we employ a novel validation protocol for social navigation scenarios to compare our gesture-aware policy with a standard MPC policy.
\section{PRIOR WORK}

Due to its importance to robotics, research on social navigation has recently garnered considerable attention, spanning a breadth of methods which we cannot comprehensively discuss. For a general overview, interested readers are referred to recent survey papers by Gao and  Huang~\cite{10.3389/frobt.2021.721317}, Charalampous et al.~\cite{CHARALAMPOUS201785}, Mavrogiannis et al.~\cite{mavrogiannis2021core}, Kruse et al.~\cite{KRUSE20131726}, Rios-Martinez~\cite{Rios-Martinez2015}, and Mirsky et al.~\cite{mirsky2021prevention}. The following discussion focuses on approaches to social navigation and human robot interaction with methods  closely related to our work. 

A variety of researchers have explored gestures for robotic control. For teleoperation, researchers used hand tracking with a Microsoft Kinect device \cite{zhang2012microsoft} ~\cite{du2012markerless}, a Leap motion ~\cite{zhang2019gesture}, or a camera to direct a robot's end effector.
For robot navigation, \cite{saha2017novel}, \cite{saha2018gesture} extracted gestures from a Kinect sensor to generate navigation commands for a teleoperated robot in a non-social setting. \cite{iocchiperson} used hidden Markov models to detect six gestures with a Kinect sensor and mapped them to hand-crafted robot navigation behaviors. \cite{chaithanya2022survey} used a wearable gesture detector to control a mobile robot. Finally, \cite{calinon2007learning}, \cite{bandera2010vision} and \cite{kuniyoshi2015learning} used imitation learning to replicate human gestures on a robot platform.

Researchers have also examined gestures for social communication. \cite{hoffman2010gesture} modeled musical improvisation as a series of gestures so a robot marimba player could improvise with a human. In \cite{neto2019gesture}, researchers designed a gesture framework for humans to collaborate with a co-worker robot on an assembly task; \cite{ende2011human} approached a different collaborative task between humans and robots using a gesture framework. Other gesture-based human-robot communication systems have been developed for settings where verbal communication may be challenging, such as emergency settings \cite{de2013field} or underwater \cite{chiarella2018novel}.
\section{METHOD}

Our main goal is to develop a gesture-aware navigation policy which can guide robots around humans in a socially compliant manner while responding to provided gestures. We define our \textit{Gesture2Path} policy by combining a sequential neural network as a high-level planner with MPC for low level control (Fig.~\ref{fig:pipeline}). The high-level planner identifies humans and the gestures they perform from images, while the low-level controller safely navigates the robot from one waypoint to another. The advantage of this setup is that the high-level planner and the low-level controller complement each other. MPC performs well for navigating from one waypoint to another, even around dynamic obstacles, based on point cloud data obtained from the environment. However, MPC fails to detect humans and their gestures as it does not use RGB data. Conversely, the high-level planning neural network is able to detect humans and their gestures from sequences of images, but fails to reliably predict waypoints for the navigation task.

\subsection{Gesture-aware Social Navigation}

To control a robot in response to a human and their gestures, we employ a sequential neural network~\cite{sutskever2014sequence} that, given a history of observations, outputs a sequence of future positions. We learn this model from demonstrations of successful navigation trajectories of human-robot interactions and show that the network is able to learn to interpret human gestures.

In more detail, we denote by $s=(I, O, p)$ the robot state consisting of an RGB image $I$ from a headmounted camera, an occupancy grid $O$ from a LiDAR sensor, and a robot position $p$ in the world coordinate frame. A demonstration trajectory of length $n$ can be described as a sequence of states: $(s_1, \dots, s_n)$. An occupancy map is defined as the 2D projection of a point cloud obtained from a LiDAR scanner. 

Given a history of states, we aim to compute controls that would result in the future robot positions as encoded by the demonstrations. In particular, at step $n$ we utilize a history of $k$ states from the demonstration to predict $l$ future positions:
\begin{equation}
    (p_{n+1}, ..., p_{n+l}) = \textrm{Gesture2Path}(s_n, \dots, s_{n-k}, g), 
\end{equation}
where $g$ is the goal position of the robot.

In theory, we could use a neural network to directly learn robot control commands. Several approaches do learn direct mappings from pixels to motor controls~\cite{NIPS2003_b427426b,1307456,PETERS2008682}. However, learning an end-to-end robot control policy remains challenging, as methods tend to require large training datasets and easily overfit to the underlying robot dynamics. 

Therefore, we employ MPC as a low level controller that, given an intermediate goal position, drives the robot to this position while avoiding collisions. The MPC algorithm takes care of converting target positions to low level torque controls.

To navigate a robot a MPC controller requires the intermediate next position encoded in the robot frame. Thus, we need to convert the demonstrated robot positions from world frame to a sequence of egocentric positions, each in the frame of the previous position. This can be done by calculating the deltas between subsequent positions: $\delta_i = p_i-p_{i-1}$ for $i=1, \dots, l$, which leads to the final model formulation:
\begin{equation}
    (\delta_{n+1}, \dots, \delta_{n+l}) = \textrm{Gesture2Path}(s_n, \dots, s_{n-k}, g).
\end{equation}

\subsubsection{Sequential Model Architecture}
\label{sec:model_architecture}
\begin{figure}
\centering
\includegraphics[width=\columnwidth]{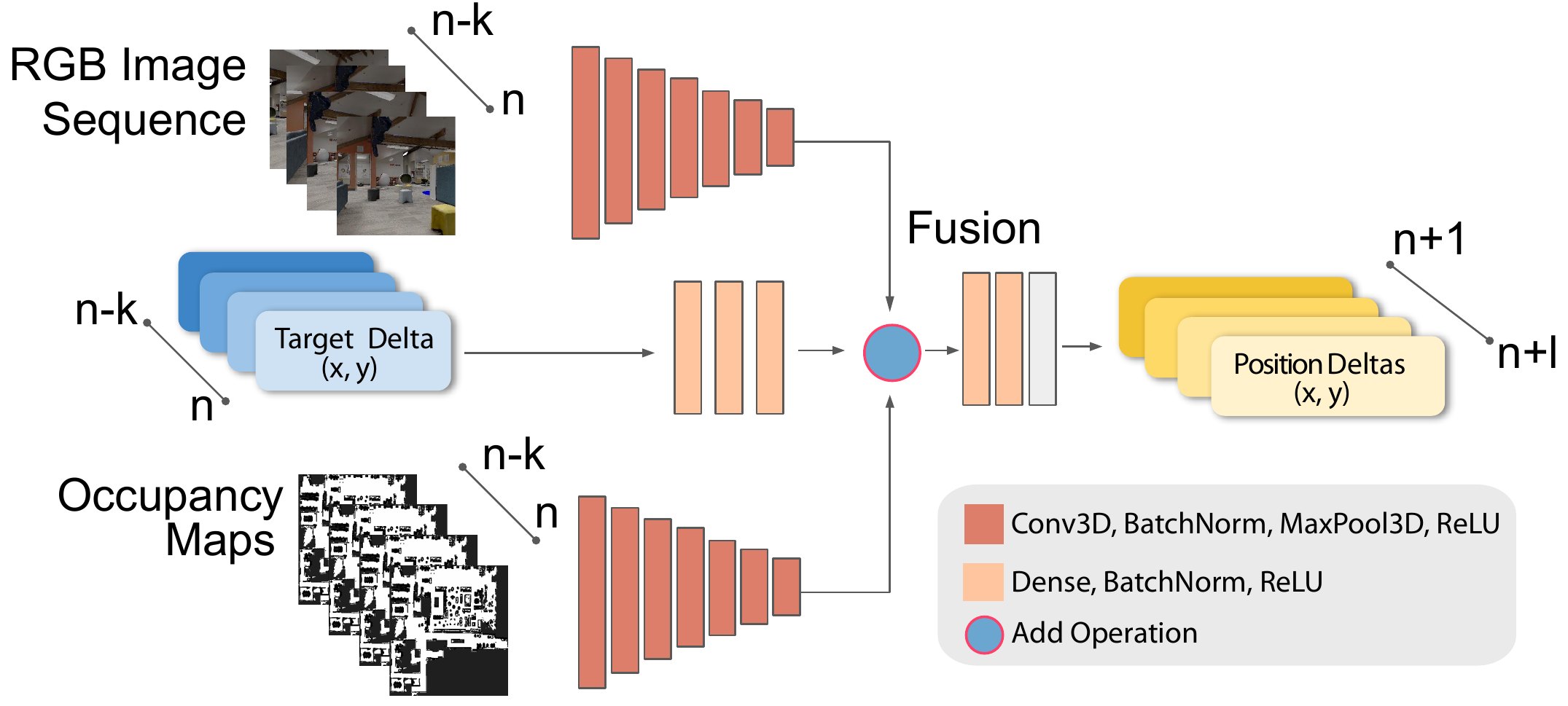}
\caption{\small{We use a 3D CovNet neural network architecture that joinly embeds sequences of RGB images $I$, occupancy maps $O$, and target deltas $\tau$ into a fused embedding space. The network is trained so as to predict sequences of position deltas that we use to generate an intermediate goal position for an MPC algorithm.}}
\label{fig:network_architecture}
\end{figure}

We train the sequential model by simultaneously embedding subsequences of RGB images $I$, the occupancy maps $O$, and the target deltas $\tau = g - p_n$ into a joined latent space. Specifically, our neural network architecture uses 3D convolutional layers to obtain two 64-dimensional embeddings from RGB images and occupancy maps of the input subsequence with a resolution of 256x256 pixels. To combine the target deltas ($\tau$) with the image embeddings we also project them into a 64-dimensional latent space. We then add the embeddings and pass them to three dense layers to obtain a sequence of $l$ position deltas~($\delta$). For the last dense layer we use a linear activation for regression. Fig.~\ref{fig:network_architecture} shows this network architecture. Compared to other approaches for action recognition (e.g.~\cite{https://doi.org/10.48550/arxiv.2201.08377}), the goal for defining a lightweight multi-modal architecture for detecting gestures is to ensure small inference times ($<$ 30ms).

While our policy is agnostic to the choice of the sequential model, we found  a common 3D-convolutional neural network similar to \cite{Carreira2017} provided the best performance for our experiments. We train the network via imitation learning using expert sequences of human-robot interactions captured by manually driving the robot with a human operator (see Section~\ref{sec:data_collection}). 

\begin{figure}
\centering
\includegraphics[width=\columnwidth]{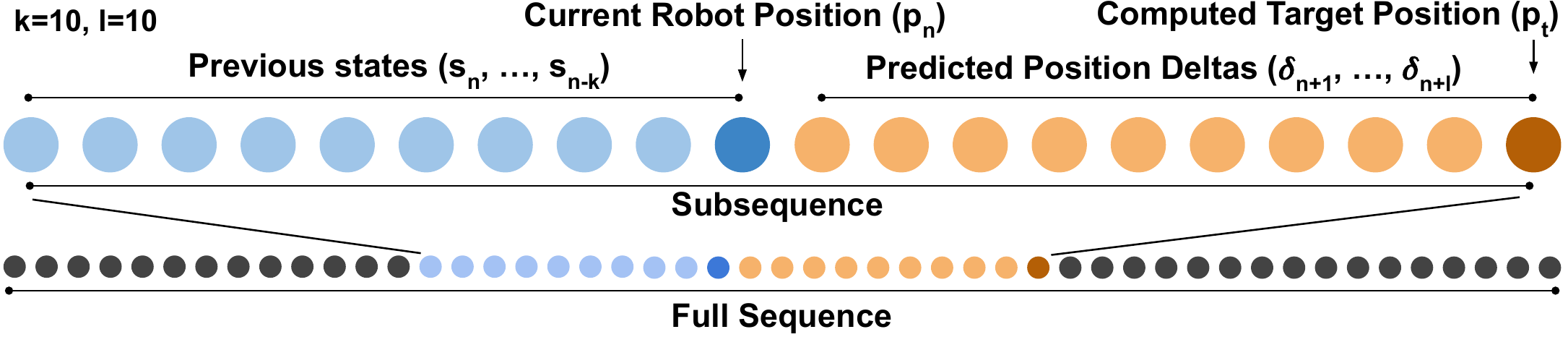}
\vspace{-1mm}
\caption{\small{Our sequential neural network predicts a subsequence of position deltas ($\delta$)  from a history of past states $s$. We use the predicted position deltas to compute an intermediate target position~($p_t$).}}
\vspace{-1mm}
\label{fig:sequence}
\end{figure}

Fig.~\ref{fig:sequence} illustrates the parameters of our learning setup. We train our sequential neural network on subsequences of the collected expert trajectories (full sequence). The training objective is to predict a sequence of position deltas (with $l$ steps into the future) based on a history of $k$ previous states. The goal is to use the predicted position deltas to compute an intermediate target position ($p_t$) from the current position ($p_n$). To train the model we sample all possible subsequences from our dataset of expert trajectories. 


\subsubsection{Action Parameterization}
We use MPC for low-level robot control. Specifically, we formulate the goal following as an optimization problem and use an iterative linear quadratic regulator (iLQR) solver, Trajax \cite{trajax2021github}, that minimizes a hand-engineered cost function, similar to \cite{6907001,8371312,8594448}. The optimization problem was solved at each control step and the predicted action at the first step (i.e. velocity command) is sent to the robot. The inputs to MPC include the occupancy grid and relative goal position; its cost function combines time-weighted goal penalties, margin-offset collision penalties, and a weighted control penalty; and its action space is linear and angular velocities. This MPC policy provides smooth navigation and fast reaction times for differential drive robots and is superior in our tests to a re-implementation of the well-performing reinforcement learning policy in \cite{francis2020long}.

We use MPC in two ways: first we establish a baseline, where we use the algorithm to navigate from a start to a goal position without additional visual inputs. Second, we use MPC for our gesture-aware policy by tracking waypoints that are generated by the sequential neural network. In both cases, the action space are linear and angular velocities.

\begin{figure}[t]
\centering
\includegraphics[width=\columnwidth]{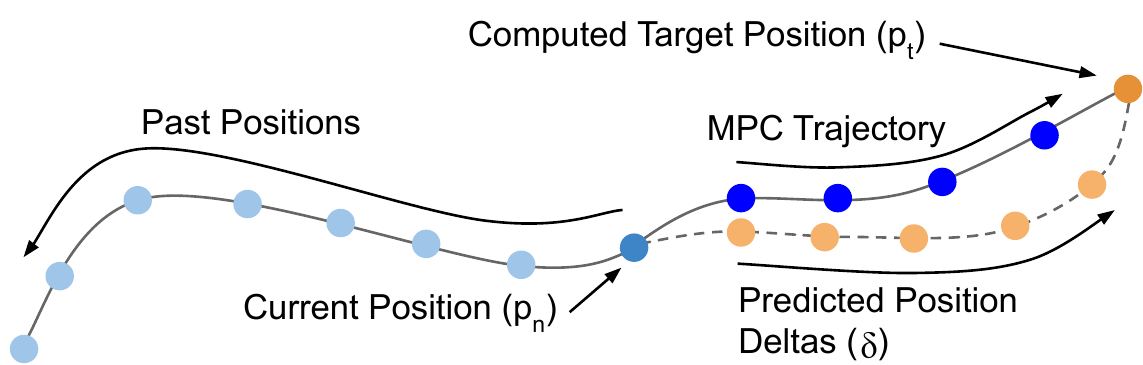}
\caption{\small{Once the intermediate target position $p_t$ has been computed from the predicted position deltas, we use MPC to compute an optimal trajectory from the current position $p_n$ to the target position. Note that the trajectory represented by the predicted position deltas and the computed MPC trajectory may diverge -- we do not want to over-constrain the MPC algorithm to find an optimal trajectory.}}
\label{fig:il_mpc}
\vspace{2mm}
\centering
\includegraphics[width=\columnwidth]{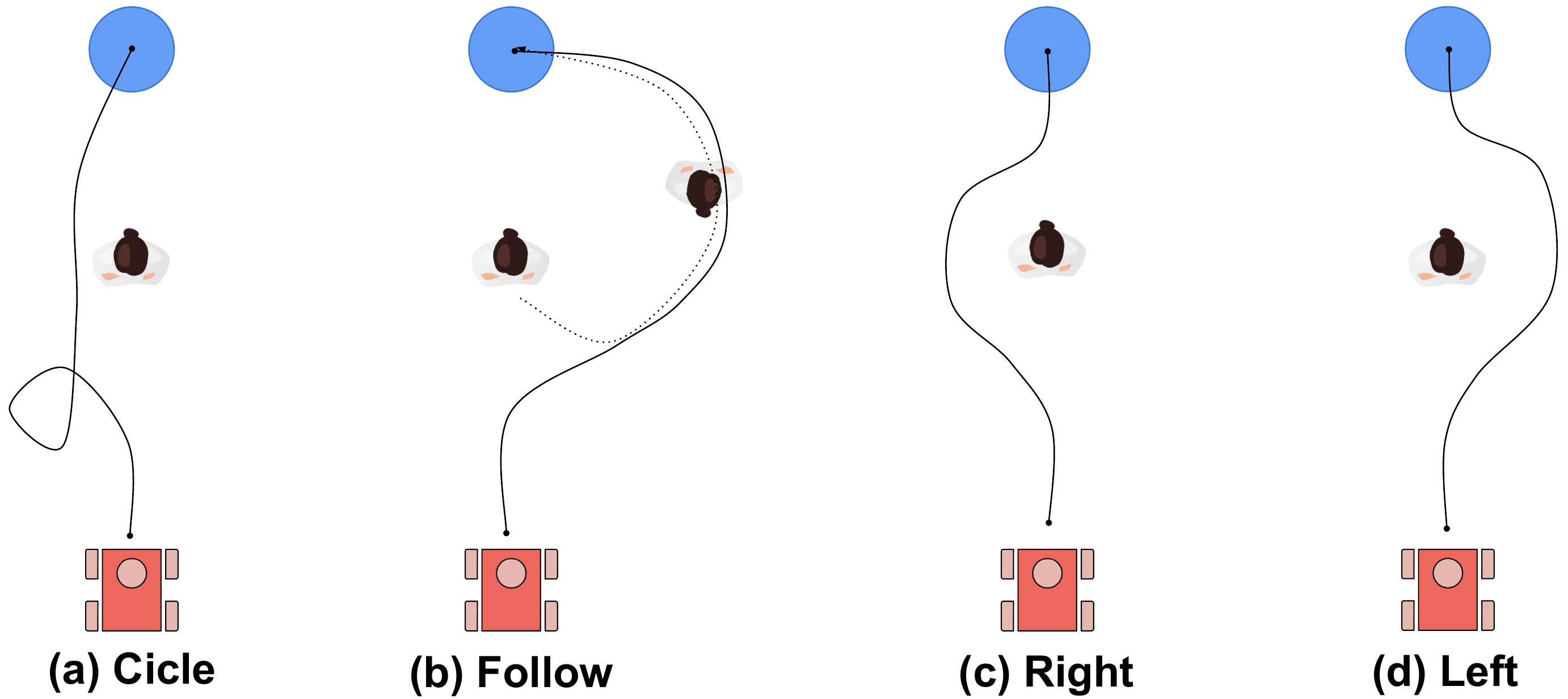}
\vspace{-6mm}
\caption{\small{We test \textit{Gesture2Path} with three policies to respond to four gestures: circle (a), follow (b), right (c), and left (d). For each scenario we only define the gesture (e.g.~pointing with the hand to the left), the corresponding expected robot behavior (e.g.~drive left), and the start and end positions of the robot trajectory. The robot is tasked to navigate from its start position to a goal area (blue circle). A person is standing in the center of the room and interacts with the robot by performing different gestures to initiate the robots' response.}}
\label{fig:gesture_scenarios}
\vspace{-4mm}
\end{figure}

For gesture-aware navigation, we use the sequential neural network described in Section~\ref{sec:model_architecture} to predict position deltas from a sequence of previous states. From the position deltas ($\delta$) we compute an intermediate target position $p_t = p_n + \sum_{i=n+1}^{i=n+l} \delta_i$ for MPC based on the current position~$p_n$. The MPC algorithm then computes an optimal path from the current position $p_n$ to the intermediate target position~$p_t$. 

At every time step we collect the robot state $s_n$. To generate gesture-aware trajectories we use the neural network to make predictions every $k$ time steps (Fig.~\ref{fig:pipeline}, Planning Loop), while the MPC algorithm runs at a frequency of 10 Hz (Fig.~\ref{fig:pipeline}, Control Loop). With $k=10$, this means that the neural network is used to predict a new intermediate target position every 10 time steps, while the MPC algorithm continuously generates control commands to drive toward the intermediate target position. Note that the number of previous states (defined by $k$) used to make predictions with the neural network need not be the same as the number of predicted position deltas (defined by $l$). The neural network may be triggered to predict a new intermediate target position before the robot reaches the previously computed one. This ensures that the gesture-aware policy is susceptible to chances in the observed gestures.  

Moreover, as illustrated in Fig.~\ref{fig:il_mpc}, the trajectory represented by the predicted position deltas of the sequential neural network and the MPC-computed trajectory may diverge. While we want the robot to closely follow the predicted position deltas to implement gesture-aware behavior, we do not want to over-constrain MPC to find an optimal trajectory from the current position to the intermediate target position. The planning loop based on the sequential neural network provides general direction while considering observed gestures, and the control loop based on MPC ensures safe navigation.

\section{EXPERIMENT SETUP}

The robots we use are 7 DOF mobile manipulators with a single arm and a rectangular base. Each robot is equipped with a head-mounted camera and a LiDAR sensor. Robots also have access to an existing static map of their environment to assist with localization and point-to-point navigation. 
\subsection{Gesture Scenarios}
To test the capabilities of \textit{Gesture2Path}, we use the protocol proposed in~\cite{Pirk2022SocialNavProtocol} to define the four gesture scenarios (illustrated in Fig.~\ref{fig:gesture_scenarios}): Circle, Follow, and Left/Right
\begin{itemize}
\item \textbf{Circle}: In the make a circle scenario, the demonstrator would stand in the center of the room and raise their right hand straight up to the ceiling. The robot operator then drives the robot to make a counterclockwise circle before passing the human on their left (Fig.~\ref{fig:gesture_scenarios} a). 
\item \textbf{Follow}: In the follow me scenario, the demonstrator would lift both hands out to the side while keeping their feet together, making a ``T'' shape. After holding this shape for 2 seconds, they would turn over their left shoulder and walk along a wide arc prior to passing through the end position region. Once the gesture has been observed the robot operator would then drive the robot so as to follow the demonstrator  (Fig.~\ref{fig:gesture_scenarios} b). 
\item \textbf{Left/Right}: In the left/right scenario, the demonstrator would stand in the center of the room and put their arm out to the left or right while leaning in that direction. The robot operator would then drive the robot to pass along the side of the human's outstretched arm (Fig.~\ref{fig:gesture_scenarios} c, d). 
\end{itemize}

We selected these gestures based on the following criteria for the human and the robot: (1) the robot solves a navigation tast beginning at one side of the room and ending at the other; (2) the robot would move continuously from beginning to end without pausing or stopping; (3) the gesture instructions would be simple and easy to teach the experiment participants; (4) the gestures would correspond to clear changes in robot behavior; (5) the gestures require the robot to react to a single participant. While the gestures used for the Circle and Follow scenarios are distinct, the Left and Right gestures represent more subtle variations for the sequential model to disentangle. 

\subsection{Data Collection and Training}
\label{sec:data_collection}

\begin{figure}[t]
\centering
\includegraphics[width=\columnwidth]{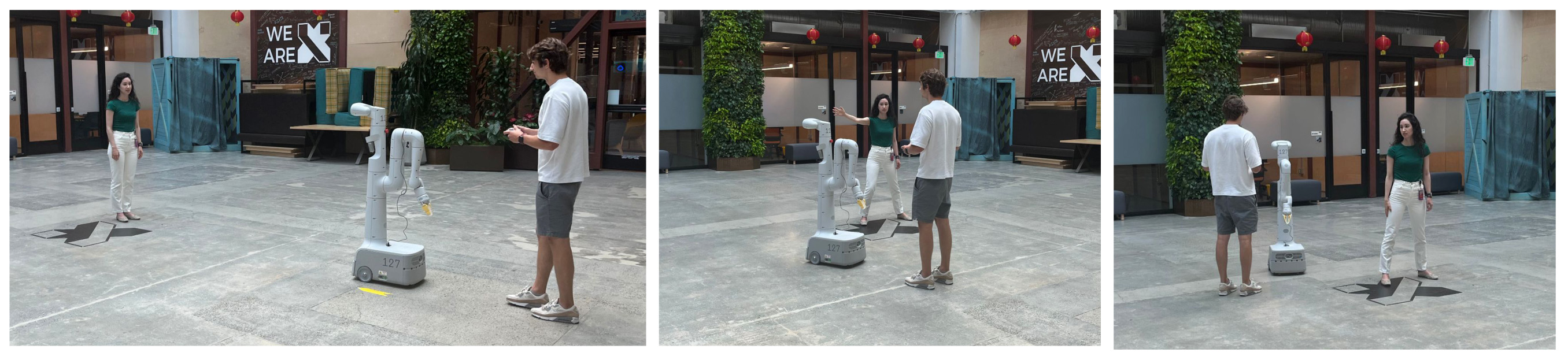}
\vspace{-6mm}
\caption{\small{For data collection the robot is controlled by a joystick handled by an experienced teleoperator. The human demonstrator stands in the center of the room and performs the gesture as well as the next sequence of human motions (if in the follow condition). The teleoperator moves the robot from the beginning position to the ending position with the correct response.}}
\vspace{-4mm}
\label{collection}
\end{figure}

To train our \textit{Gesture2Path} policy we collected a dataset of 277 trajectories (40 for follow, 30 for circle, 207 left and right). Each trajectory represents a 20-40 second sequence that includes a human-robot interaction. The human performs one of the four defined gestures, while the robot solves a point-to-point navigation task from a start to a goal position. All training data was collected in the real world in a large, open atrium space over several different days and times (Fig.~\ref{collection}). 

To collect our dataset we used the following procedure: an expert robot operator would connect to the real robot using a Logitech F710 gamepad. The robot operator would stand behind the robot and follow it while driving it forward. The gesture demonstrator would begin by standing in the middle of the room. Once the teleoperated robot started to move, the gesture demonstrator would perform the gesture for 2-3 seconds. Depending on the gesture scenario, the demonstrator would either stay in place or move appropriately. 

We split the collected trajectories into 90\% training and 10\% validation data.  We then trained our sequential neural network on subsequences ($k=10$, $l=10$, for most of our experiments) of RGB images and occupancy maps. We used the Adam optimizer with a standard $l2$ loss and selected checkpoints of the network for our policies based on the lowest $l2$ error measured on the validation dataset. The training of our network usually converges after 120 epochs, which usually takes 6 hours of training on a single V100~GPU. 

\vspace{-2mm}
\subsection{Experiment Design}

To validate the effectiveness of our gesture-aware policies we conducted a user-study to obtain in-situ experience ratings. In the following, we discuss the setup of our experiment for our \textit{Gesture2Path} polices as well as for the MPC baseline. 

\subsubsection{User Study Design}
We designed a within-subjects experiment where participants performed the four gestures while the robot was running either \textit{Gesture2Path} or a baseline of the unmodified MPC policy. Note this baseline MPC policy is itself a highly performant navigation policy, exceeding in our internal tests a reimplementation of \cite{francis2020long}. The study occurred in the same location where the training data was collected. 

Participants arrived at the location and filled out an introductory survey asking about their dominant hand and number of months they had been working near these robots. The participants then performed each of the four gestures a total of 70 different times in the center of the room. Each the trials consisted of the robot crossing from one side of the room to the other; the human participants would perform the gesture in the center of the room (and during the follow scenario, walking to the edge of the room). Once the trial was completed the participant would immediately answer four questions about the robot's performance. The questionnaire was defined based on a 5-level Likert scale. Each of the four questions were rated on a scale of 1-5, 1 being \textit{Strongly Disagree} and 5 being \textit{Strongly Agree}. Participants had the option to answer 0 if the question did not apply. The questions were:
\begin{enumerate}
    \item The robot maintained a safe distance at all times. (Safe)
    \item It was clear what the robot wanted to do. (Clarity)
    \item The robot responded correctly to my gesture. (Correct)
    \item The robot paid attention to what I was doing. (Trust)
\end{enumerate}

We ran the gesture trials 10 times for each of our \textit{Gesture2path} policies (Follow, Circle, Left/Right). For the MPC baseline we ran 10 trials for the Follow and Circle scenarios and 5 trials for the Left and Right scenarios. 

We randomized policy ordering for each participant. For example, one participant may have completed MPC Left/Right, \textit{Gesture2Path} Left/Right, MPC Follow, Policy Follow, MPC Circle, Follow Circle. After the gesture trials, the participants filled out a closing survey asking about their age, gender, cultural group, height, highest degree of education received, primary field of work or study, as well as their experience with STEM, robotics in general, and specifically the used robot. The experiment took approx. 90 minutes to complete.

\subsubsection{Participants}
12 individuals volunteered to participate in the user study, including 2 women and 10 men. 8 participants identified as Asian, 2~as White, 1~as Hispanic/Latino, and 1~Middle Eastern. 3 participants were ages 18-24, 8 were ages 25-34, and 1 declined to state. All participants listed their \textit{right hand} as the dominant hand and all participants had previously used the robot in their work.

\begin{figure}[t]
\centering
\includegraphics[width=.8\columnwidth]{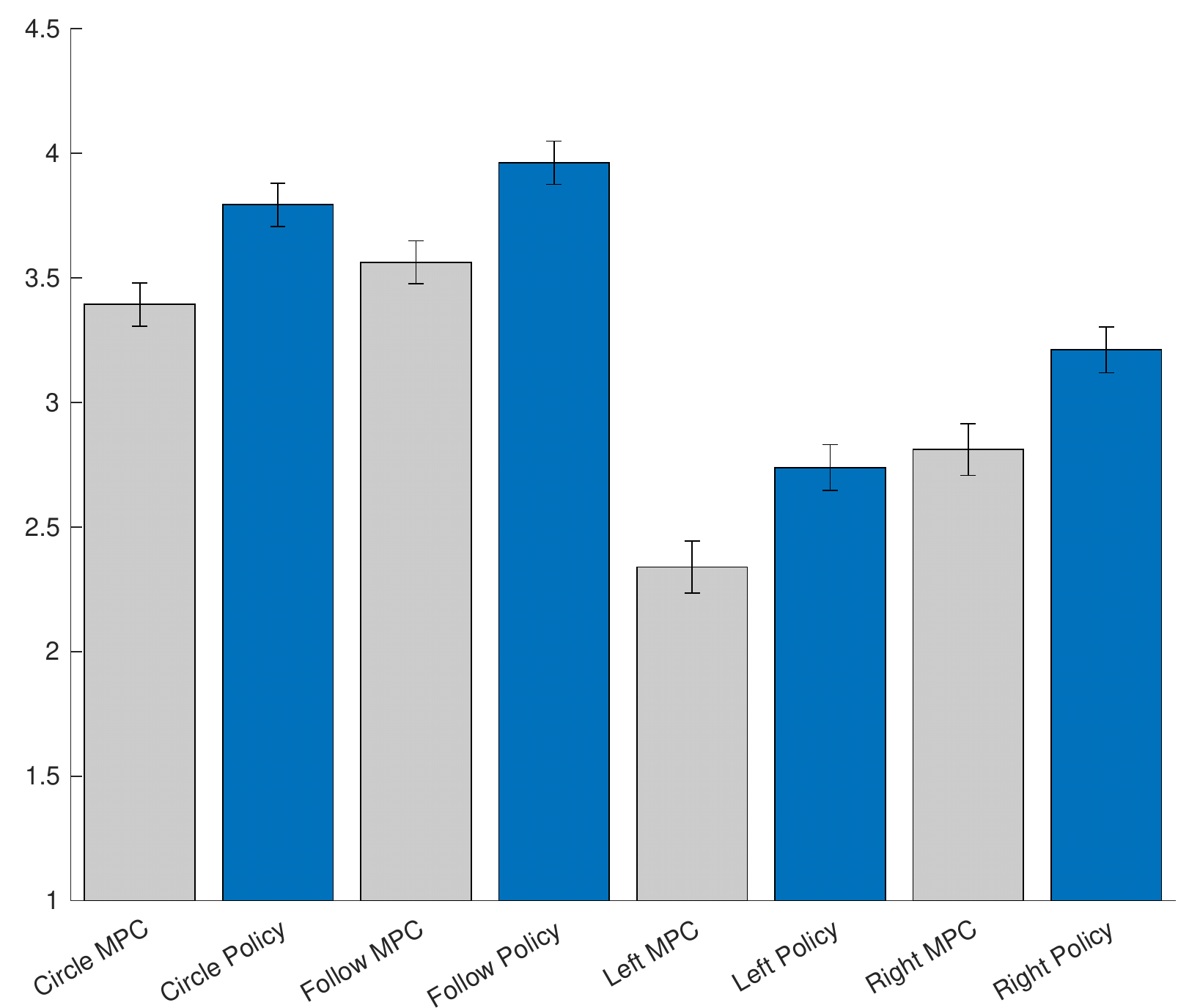}
\caption{\small{The means and standard errors for all four questions for policy and gesture condition. The MPC conditions are in grey and \textit{Gesture2Path} policy in blue. Participants rated the question on a scale of 1 Strongly Disagree to 5 Strongly Agree. Zero was reserved for Not Applicable. The policy rated overall more highly for each of the four questions when compared to MPC. A paired t-test was performed on each MPC/policy bar for each gesture (i.e. comparing circle MPC to circle policy). All of these tests were statistically significant with a p-value~$<$~.01. We can conclude that the policy performs overall better than MPC for each of these gesture scenarios.}}
\vspace{-3mm}
\label{fig:allmeans}
\end{figure}
\section{RESULTS}

\begin{figure}[t]
\centering
\includegraphics[width=\columnwidth]{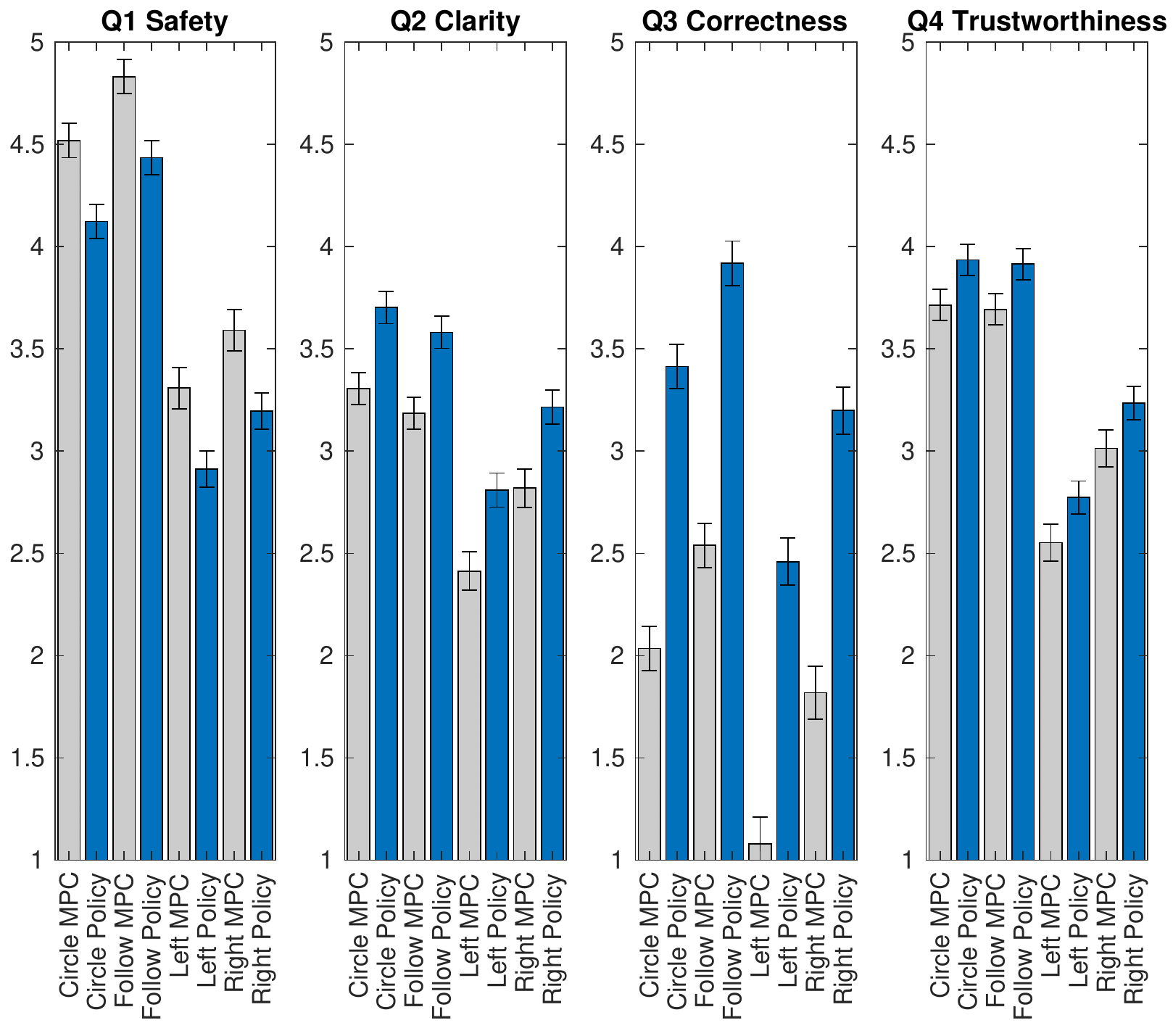}
\vspace{-6mm}
\caption{\small{All averages and standard errors for each question for each gesture, \textit{Gesture2Path} is in blue, MPC is in grey. Question 3 (Correctness) shows the largest difference between \textit{Gesture2Path} and MPC. A repeated measures ANOVA on Question 3 showed a statistically significant difference.}}
\vspace{-5mm}
\label{fig:16bars}
\end{figure}

All participant ratings for each gesture were grouped together by \textit{Gesture2Path} and MPC (i.e. questions 1-4 for Circle gesture, \textit{Gesture2Path}, were all grouped). These grouped responses were averaged, as shown in Fig.~\ref{fig:16bars}. A paired t-test was performed for each gesture vs policy group (i.e. Questions 1-4 for Circle \textit{Gesture2Path} vs Questions 1-4 for Circle MPC) and all differences are statistically significant with a p-value~$< .01$. Therefore, we conclude that the \textit{Gesture2Path} policy overall performs better than MPC for all four gesture scenarios.

To determine which questions had the strongest effect on \textit{Gesture2Path's} improvement over MPC, we ran a repeated measures ANOVA on each question. The repeated measures ANOVA comparing the effect of \textit{Gesture2Path} vs MPC on Question 3 (Correctness) showed a statistically significant difference: F(within groups df) = 115.563, p-value = .059.

Separate repeated measures ANOVAs were performed to compare the effect of \textit{Gesture2Path} vs MPC on Question 1 (Safety), Question 2 (Clarity) and Questions 4 (Trustworthiness). Fig.~\ref{fig:allmeans} shows that \textit{Gesture2Path} (blue bars) is higher in all cases for Question 2 (Clarity) and Question 4 (Trustworthiness), while MPC (grey bars) is higher for Question 1 (Safety). However, for Questions 1, 2, and 4, the ANOVA did not show statistical significance. Therefore, the critical improvement that elevated \textit{Gesture2Path} over MPC was the policy's correct response to the human gesture (Question 3).

The \textit{Gesture2Path} policies took longer than MPC in all cases, as summarized in Table I. For Circle and Follow, this was likely due to the gesture response trajectory requiring a longer distance for the robot to travel. For the Left/Right Policy, the additional time was likely due to the wider distance needed by the gesture, as well as how quickly the human operator drove the robot during data collection. A paired t-test was performed on each Gesture2Path and MPC pairing for each gesture and the results were statistically significant.

\begin{figure*}[t]
\centering
\includegraphics[width=\linewidth]{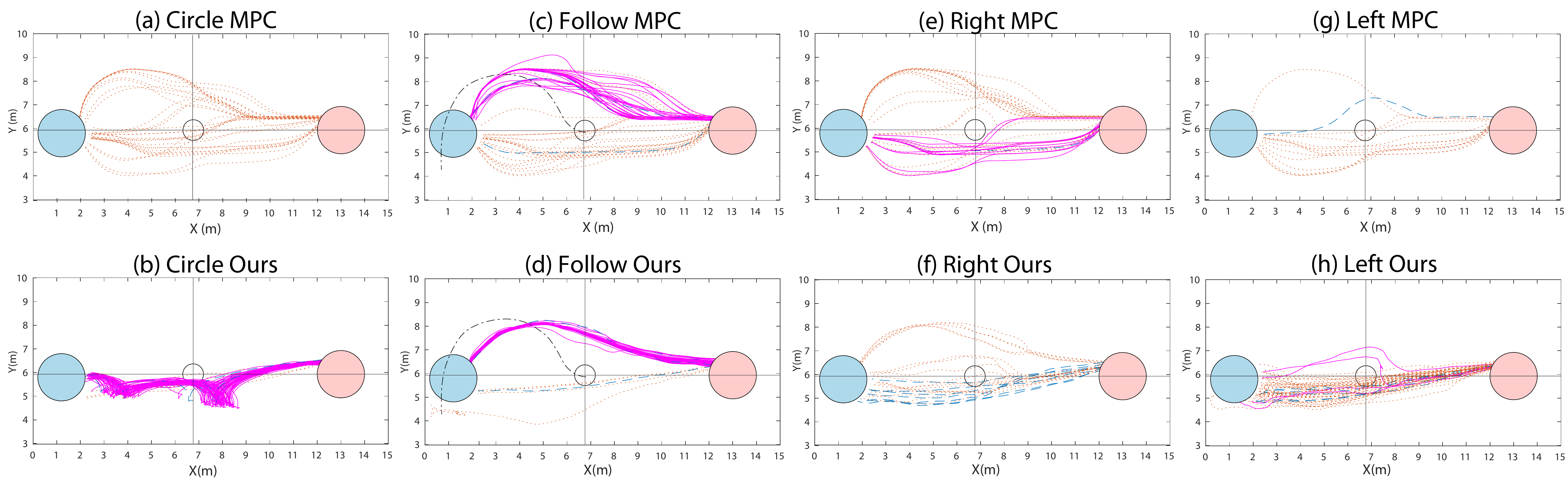}
\vspace{-3mm}
\caption{\small{2D plots of the robot trajectories and corresponding participant ratings (Question 3 only) for all four gestures with \textit{Gesture2Path} (Ours) and MPC. The robot starts in the red circle and is tasked to navigate to the blue dot. The human is the black circle in the center. For the Follow plots, there is an additional black alternating dot-dash line to indicate the rough path that the human walks after they make the gesture in the center of the room. Favorable participant ratings (4 or 5 on a 5 point scale) are solid lines in magenta. Neutral participant ratings (3 on a 5 point scale) are the dashed line in blue. Negative participant ratings (1 or 2 on a 5 point scale) are the dotted lines in red. The MPC trajectories in the first row are all similar regardless of the gesture.  The \textit{Gesture2Path} policy robot trajectories in b) Circle and d) Follow are distinctive, correct, and consistent, with participant ratings capturing this effect. The \textit{Gesture2Path} policy does not perform as well for Right and Left sides, also reflected in the participant ratings.}}
\label{circle2d}
\vspace{-1mm}
\end{figure*}

In Fig.~\ref{circle2d} we show 2-dimensional plots of the robot navigating from the start to end positions. The positive participant ratings (4 or 5 out of 5) for Question 3 (Correctness) are shown in solid magenta. The neutral rating (3 out of 5) is show in dotted black. The negative ratings (1 or 2 out of 5) are shown in dotted red. These plots demonstrate that the participants have a clear understanding of the robot's correct response as the higher ratings correspond with the robot's correct behavior.

\begin{figure}[t]
\centering
\includegraphics[width=\columnwidth]{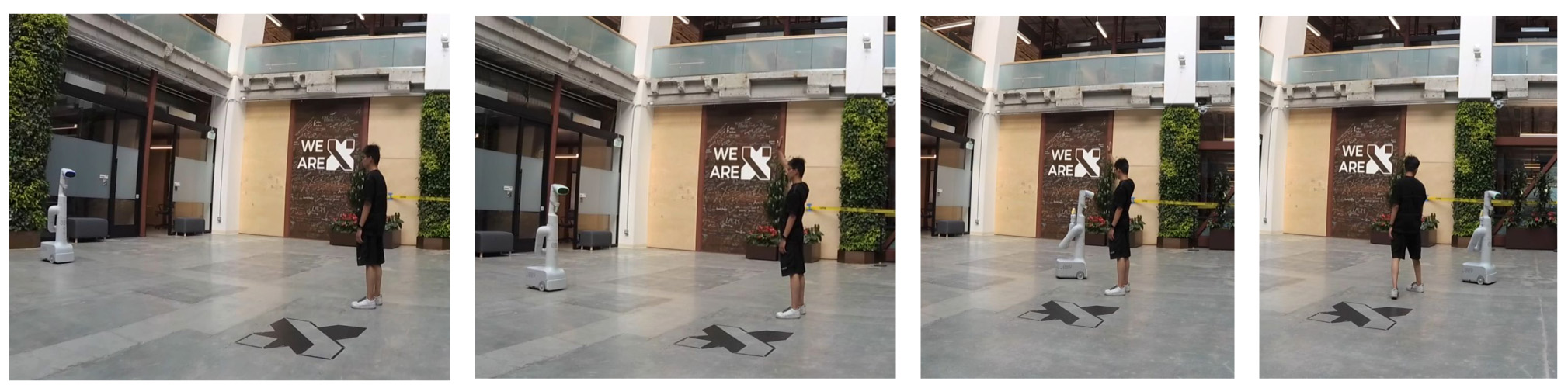}
\vspace{-5mm}
\caption{\small{Example of the MPC policy: As no image sequences are used, the policy is not able to generate trajectories in response to gestures. While the performed gesture is supposed to make the robot drive a circle, it only passes the human participant on their right side.}}
\label{fig:mpcfailure}
\vspace{-2mm}
\end{figure}

\begin{figure*}[ht!]
\centering
\includegraphics[width=\linewidth]{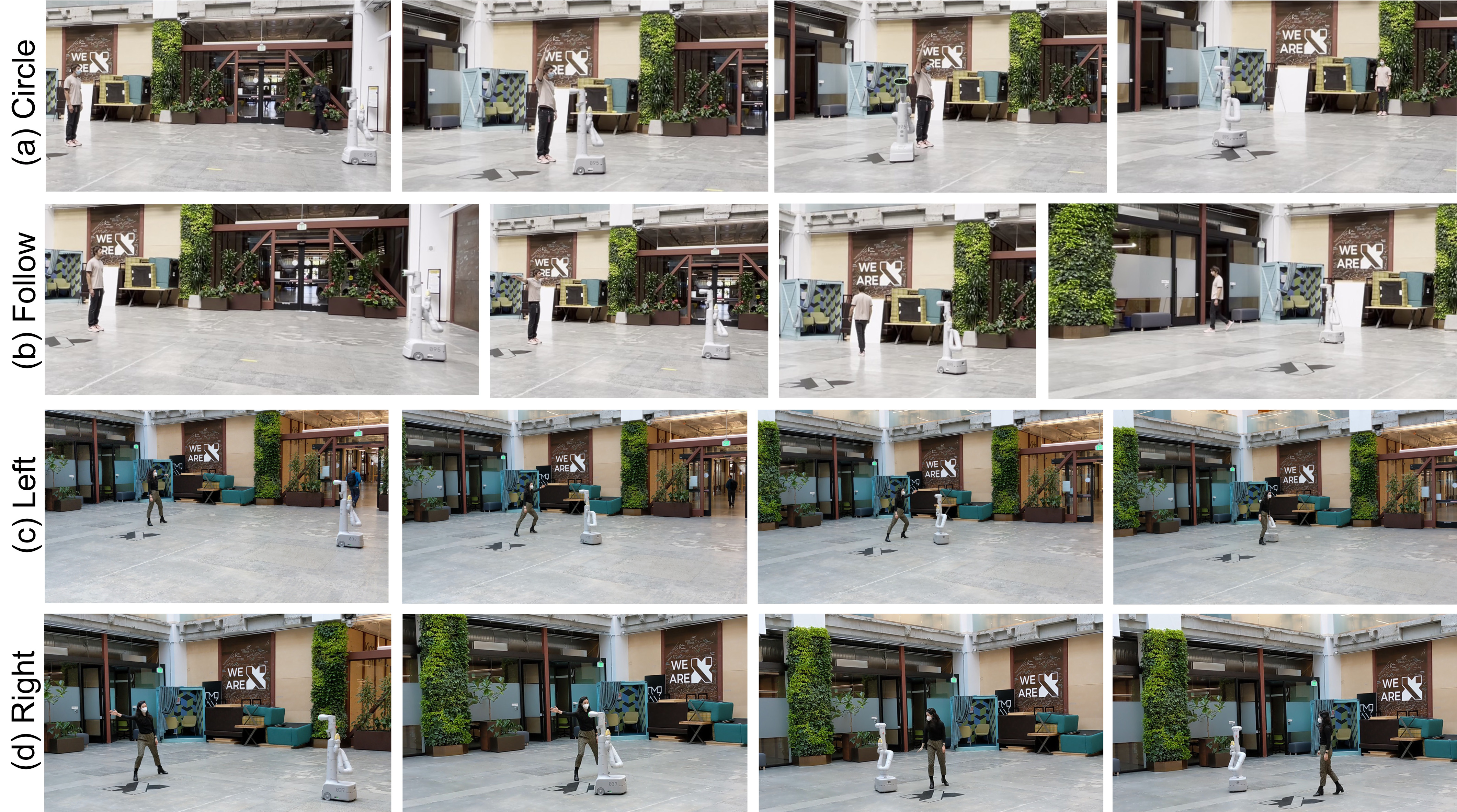}
\vspace{-1mm}
\caption{\small{The \textit{Gesture2Path} policy employed for the four gesture-aware navigation scenarios: circle (a), follow (b), left (c), and right (d). For all scenarios the robots starts on the right side of the room. The robot in the photo is from Everyday Robots.}}
\vspace{-2mm}
\label{fig:finalsixteen}
\end{figure*}

Finally, in Fig.~\ref{fig:finalsixteen}, we show examples for the four scenarios. For each policy the image sequences show (from left to right) the robot at its start location, the human performing the gesture, the gesture response of the robot, and the navigation to the goal location. As shown, our \textit{Gesture2Path} policy is able to generate socially compliant trajectories in response to the performed gestures. In contrast, as the MPC policy is not using image sequences as input to detect gestures, it is not able to generate trajectories based on the performed gestures -- an example is shown in Fig.~\ref{fig:mpcfailure}.

\begin{table}[t]
\vspace{1mm}
\captionsetup[table]{labelfont=bf, labelsep=period}
\label{table}
\begin{center}
\caption{p-values for gesture, MPC vs. policy, and relevant question using a paired t-test}
\begin{tabular}{|c|c|c|c|}
\hline
\textbf{Gesture} & \textbf{Gesture2Path, MPC} & \textbf{Avg duration (sec)} & \textbf{p-value}\\
\hline
Circle & Gesture2Path, MPC  & 56, 35 & $<.001$\\
\hline
Follow & Gesture2Path, MPC & 45, 38  & $<.01$\\
\hline
Left & Gesture2Path, MPC & 45, 38  & $<.01$\\
\hline
Right & Gesture2Path, MPC & 46, 33  & $<.01$\\
\hline
\end{tabular}
\end{center}
\emph{}The Gesture2Path policy took significantly longer than MPC in all cases, likely due to the increased distance requirements of the correct response path or the speed of the demonstrations.
\vspace{-4mm}
\end{table}

\subsection{Discussion and Limitations}
Our MPC baseline can reliably navigate around obstacles with success comparable to published state of the art \cite{francis2020long}. In this work we show that using imitation learning in tandem with MPC simultaneously allows us to maintain this performance level while enabling gesture-aware navigation. The aim of the study was to show that robots can respond correctly to human gestures and we observe this in the results.

The trained \textit{Gesture2Path} policies perform better than baseline MPC policy for all gestures in the participant ratings (Fig.~\ref{fig:16bars}). Question 3 (Correctness) was statistically significant, showing that \textit{Gesture2Path} is able to correctly identify and respond to human gestures. The difference in ratings between baseline MPC and \textit{Gesture2Path} was larger for the Circle and Follow policy than for the Right and Left policy, meaning that either participants were more certain of the robot's correct response or that  \textit{Gesture2Path}'s behavior was more distinctive than baseline MPC. Baseline MPC performed better than \textit{Gesture2Path} for all four gestures on Question 1 (Safety). The higher safety rating follows from baseline MPC's tendency to give humans a large amount of space to avoid collisions.

Recall we trained three policies: 1) Circle only, 2) Follow only, and 3) Left and Right. The third policy must generate different paths for two gestures. As baseline MPC tends to go right when it encounters obstacles, it appears correct to participants half the time (high clarity rating, Question~2). 

We limited the number of questions to four due to practicality reasons; running the study became quite lengthy. The study ran for a minimum of 90 minutes, so participants may have been fatigued towards the end. One participant elected not to complete the study halfway through due to the time commitment. To mitigate this fatigue effect in the data, conditions were randomized for each participant. 

Furthermore, not all participants perform the gesture the same way. Despite watching the same video in order to learn the gesture, each human has different timings, range of motion, and qualitative emphasis. Therefore, it is promising that \textit{Gesture2Path} can generalize to these performance variations.
\section{CONCLUSION AND FUTURE WORK}

We have introduced \textit{Gesture2Path}, a novel social navigation policy, to enable gesture-aware point-to-point navigation. Our approach combines a novel sequential neural network which can predict position deltas from images with MPC for generating linear and angular velocities to control a robot. As the sequential neural network is trained on images of the observed gestures it can generate positions as waypoints which implement socially-acceptable and gesture-aware navigation behavior. On the other hand, MPC allows us to safely navigate from the current position of the robot to an intermediate goal position. Based on four scenarios we have shown that our \textit{Gesture2Path} policy is able to respond correctly, clearly, and legibly to a range of human gestures, while it also outperforms the baseline MPC algorithm devoid of visual inputs.  

In future work, we plan to train our policy so as to handle multiple gestures simultaneously. This step is important for deploying \textit{Gesture2Path} in a working robot environment, where the aim is to maximize the number of gestures with the smallest number of concurrently running policies. We will also explore multiple humans gesturing simultaneously. As another direction for future work we are interested in exploring a closer connection of the introduced sequential neural network and MPC. While keeping both algorithms separate has intriguing advantages from an engineering point of view, it seems likely that a closer integration of understanding gestures and low-level control would lead to better performance. Finally, we want to use our \textit{Gesture2Path} policy for learning even more complex and diverse sets of gestures to enable even more advanced robot navigation behavior.

{\small
\bibliographystyle{ieee}
\bibliography{main}
}

\vfill

\end{document}